\documentclass[letterpaper]{article} 
\usepackage{aaai23_arxiv}  
\usepackage{times}  
\usepackage{helvet}  
\usepackage{courier}  
\usepackage[hyphens]{url}  
\usepackage{graphicx} 
\usepackage{natbib}  
\usepackage{caption} 
\newcommand{\ie}{\textit{i}.\textit{e}., }

\usepackage{multirow}
\usepackage{threeparttable}
\usepackage{booktabs} 
\usepackage{colortbl}
\usepackage[misc]{ifsym}
\usepackage{lipsum}
\usepackage{algorithm}
\usepackage{algpseudocode}
\usepackage{algorithmicx}
\usepackage{amsmath,amssymb} 
\usepackage{amsfonts}
\definecolor{myy}{RGB}{126,95,0}
\definecolor{mygray}{gray}{.9}
\definecolor{bblue}{RGB}{30,80,120}
\definecolor{mygray1}{gray}{.7}
\definecolor{ggray}{RGB}{127,127,127}
\definecolor{mygreen}{RGB}{93,174,86}
\usepackage{newfloat}
\usepackage{listings}
\DeclareCaptionStyle{ruled}{labelfont=normalfont,labelsep=colon,strut=off} 
\floatstyle{ruled}
\newfloat{listing}{tb}{lst}{}
\floatname{listing}{Listing}

\title{Combating Mode Collapse in GANs via Manifold Entropy Estimation}
\author{
Haozhe Liu$^{1,2}$, Bing Li$^{1 \textrm{\Letter}}$, Haoqian Wu$^{3,4}$, Hanbang Liang$^4$, Yawen Huang$^2$, \\
Yuexiang Li $^{2 \textrm{\Letter}}$, Bernard Ghanem$^1$, Yefeng Zheng$^2$\\
}
\affiliations{
$^1$ AI Initiative, King Abdullah University of Science and Technology (KAUST), \\
$^2$Jarvis Lab, Tencent,  $^3$YouTu Lab, Tencent, $^4$Shenzhen University \\
{\small \it \{haozhe.liu;bing.li;bernard.ghanem\}@kaust.edu.sa; } {\small \it lianghanbang2019@email.szu.edu.cn ; \{linuswu;vicyxli;yawenhuang;yefengzheng\}@tencent.com} \\

}
\usepackage{bibentry}
\newcommand\blfootnote[1]{%
\begingroup
\renewcommand\thefootnote{}\footnote{#1}%
\addtocounter{footnote}{-1}%
\endgroup
}
\begin{document}
\maketitle
\blfootnote{This work was done when Haozhe Liu was an intern at Jarvis Lab.}
\blfootnote{Corresponding Authors: Bing Li and Yuexiang Li.}
\begin{abstract}
Generative Adversarial Networks (GANs) have shown compelling results in various tasks and applications in recent years. However, mode collapse remains a critical problem in GANs. In this paper, we propose a novel training pipeline to address the mode collapse issue of GANs. Different from existing methods, we propose to generalize the discriminator as feature embedding and maximize the entropy of distributions in the embedding space learned by the discriminator. Specifically, two regularization terms, \ie {D}eep {L}ocal {L}inear {E}mbedding (DLLE) and {D}eep {Iso}metric feature {Map}ping (DIsoMap), are introduced to encourage the discriminator to learn the structural information embedded in the data, such that the embedding space learned by the discriminator can be well-formed. Based on the well-learned embedding space supported by the discriminator, a non-parametric entropy estimator is designed to efficiently maximize the entropy of  embedding vectors, playing as an approximation of maximizing the entropy of the generated distribution. By improving the discriminator and maximizing the distance of the most similar  samples in the embedding space, our pipeline effectively reduces the mode collapse without sacrificing the quality of generated samples. Extensive experimental results show the effectiveness of our method which outperforms the GAN baseline, MaF-GAN on CelebA (9.13 \emph{vs.} 12.43 in FID) and surpasses the recent state-of-the-art energy-based model on the ANIMEFACE dataset (2.80 \emph{vs.} 2.26 in Inception score).
\end{abstract}

\section{Introduction}
Generative Adversarial Networks (GANs) have attracted extensive attention in recent years \cite{schmidhuber1990making,schmidhuber1991possibility,schmidhuber2020generative,goodfellow2020generative}. 
Generally speaking, a GAN consists of a generator network and a discriminator network, where the generator generates samples to fool the discriminator, and the discriminator is trained to  discriminate real and generated samples.
With such adversarial learning, GANs have shown high-fidelity results in various tasks such as image inpainting \cite{yu2018generative} and photo super-resolution \cite{li2019feedback}. 
Nevertheless,  GANs suffer from mode collapse (or training instability) \cite{mangalam2021overcoming,saxena2021generative}, hindering their further development in the generative learning community and potential applications.

\begin{figure*}[!t]
    \centering
    \includegraphics[width=0.8\textwidth]{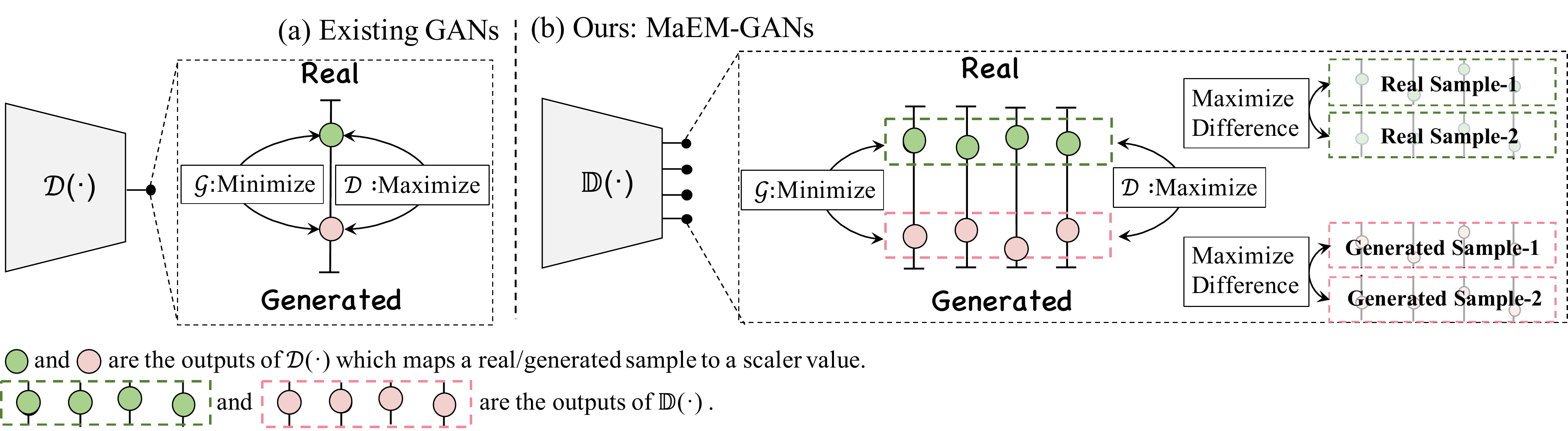}
   \caption{Illustration of our main idea for combating the mode collapse. Different from the existing GANs (a), our MaEM-GANs (b) generalize the discriminator such that it embeds an image into an $m$-dimensional space, instead of just outputting a scalar value. With the embedding vectors of real/generated samples, we maximize the entropy of distributions in the embedding space learned by our discriminator to prevent the mode collapse. }
   \label{fig:general}
\end{figure*}

To alleviate the mode collapse in GANs, many efforts have been devoted to introducing prior knowledge or adding noise on the generator side. For example, 
the conditional GANs  \cite{mirza2014conditional,odena2017conditional,brock2018large} handle the mode collapse via introducing class-level prior knowledge into GANs. By splitting the generated distribution into several sub-distributions, the complexity of image synthesis would be further reduced. As the distributions of different classes are  quite different, mode collapse can be mitigated to some extent. However,  annotating  the image classes is time-consuming and deviates from the unsupervised setting of GAN. Compared to conditional information, adding noise into networks is a more flexible solution against mode collapse. Specifically, Style-GANs \cite{karras2019style,karras2020analyzing} add Gaussian noise to the output of each convolutional layer to increase the variance of the generated samples. 
Despite the rapid progress of GAN-based approaches, mode collapse remains an unsolved and challenging problem in GANs.

Different from GANs, another kind of generative model, namely Energy-Based Models (EBMs) \cite{grathwohl2021no,geng2021bounds,lecun2006tutorial,xie2016theory,xie2017synthesizing,xie2019learning,xie2018learning}, have shown remarkable performances in circumventing the mode collapse. These methods have attracted increasing attention in recent years. Typically, EBMs estimate the density of the target data distribution and are trained via maximum likelihood. The partition function has to be estimated during the training of EBMs by adopting expensive Langevin dynamics, while the partition function is generally intractable. Consequently, since training EBMs is challenging, EMBs suffer from high computational complexity and cannot generate images with competitive fidelity.

In this paper, inspired by the success of EBMs in avoiding the mode collapse,  we propose a novel pipeline named MaEM-GANs (manifold entropy maximization)   for training GANs through bridging Wasserstein GANs (WGANs)  and EBMs.
By analyzing the connection between WGANs and EBMs,  we discover that  mode collapse in GANs can be avoided by maximizing the entropy of the generated distribution.
However, it is nontrivial
to directly estimate the entropy of the generated distribution for GANs, especially for large-scale and high-dimensional data.
Instead, since the discriminator heavily affects the training quality/stability of GANs \cite{karras2020training},  we propose to address the entropy maximization  on  the discriminator side.
In particular,  we propose to generalize the discriminator to feature embedding, such that it embeds images into a lower-dimensional embedding space (see Fig. \ref{fig:general}), different from typical GANs.  We  then maximize the entropy of distributions in the embedding space learned by the discriminator.  To further optimize such surrogate objective in an efficient and simple manner, we propose a module  named RB-MaEM based on non-parametric entropy
estimator using replay buffer.
In addition,  we introduce two regularization terms, namely,  Deep Local Linear Embedding
(DLLE) and Deep Isometric feature Mapping (DIsoMap), which encourages the discriminator to learn
the structural information embedded in the data such that  the embedding space is well formed.
Benefiting from DLLE and DIsoMap, our method, namely MaEM-GAN, maximizes the entropy in the well-learned embedding space to combat the mode collapse in GANs.
Experimental results show that the proposed MaEM-GAN outperforms the recent advanced GAN method MaF-GAN \cite{liu2021manifold} on CelebA (9.13 \emph{vs.} 12.43 in FID) and surpasses the recent state-of-the-art EBM \cite{geng2021bounds} on the ANIMEFACE dataset (2.80 \emph{vs.} 2.26 in Inception score). 

Our contributions are summarized as follows:
\begin{itemize}
\item We propose a novel training pipeline to address the mode collapse issue  in GANs, which effectively alleviates mode collapse without  sacrificing the image quality of generated images.

\item  We show that the mode collapse in GANs can be reduced  by  generalizing the discriminator as feature embedding and maximizing the entropy of distributions in the embedding space learned by the discriminator. 

\item  Extensive experiments show that our method achieves superior performances in terms of diversity and image quality on various image generation tasks, compared with  both state-of-the-art GANs and EBMs. 

\end{itemize}

\section{Related Works}

\noindent
{\bf Generative Adversarial Networks.}
To overcome the mode collapse, numerous methods have been proposed, such as introducing class-level information \cite{mirza2014conditional,odena2017conditional,brock2018large,an2019ae} or adding noises to different layers \cite{karras2019style,karras2020analyzing}. However, these methods need additional annotation, which deviates from the unsupervised setting of GAN, or introduces complex network architecture. Differently, WGANs \cite{gulrajani2017improved,arjovsky2017wasserstein} re-designed the learning objective, where the discriminator performs regression, rather than classification. By dynamically modeling the distance between the generated and real distributions, WGANs can reduce the risk for local minima and mitigate the mode collapse to a certain extent. More recently, Realness GANs \cite{xiangli2020real} and Manifold-preserved GANs \cite{liu2021manifold} generalize GANs into a high-dimensional form by mapping the output of the discriminator into a vector to enhance the realness. Since the fidelity of the given image can be judged from different views like ensemble learning, the generator can synthesize images with different attributes to fit the discriminator. The above works indicate that the discriminator is critical for GANs and can be a potential target to combat mode collapse. Motivated by these progresses \cite{zhang2019consistency,xiangli2020real,liu2021manifold} with regard to the discriminator,  we propose a new pipeline,  focusing on designing the simple yet effective constraints on the  discriminator side to prevent the mode collapse. Different from the previous methods, we explore how to maximize the entropy of distribution  in the embedding space supported by the discriminator, and how to well learn the  embedding space.  The empirical and theoretical study pointed out that the learning objective of GANs might neglect an intractable entropy term for maximum likelihood, which plays a key point in the mode collapse and can be effectively tackled by high-dimensional GANs. Manifold-preserved GAN \cite{liu2021manifold} is the most similar work to ours; however, it neglects the entropy term.

\noindent
{\bf Energy-based Models.}
There has been a rich history for EBMs, which can be traced back to Hopfield networks \cite{hopfield1982neural} and Boltzmann machines \cite{hinton1983optimal}. However, learning an EBM is difficult, since the partition function (\textit{a.k.a.} normalization constant) is intractable and hard to estimate \cite{geng2021bounds}. A common solution is based on expensive Markov Chain Monte Carlo (MCMC) sampling by generally adopting Langevin dynamics \cite{du2019implicit,xie2018cooperative} and Gibbs sampling \cite{carlo2004markov,de2018minibatch} to estimate the partition function. Some strict requirements, including parameter tuning \cite{grathwohl2019your}, early stopping of MCMC \cite{nijkamp2019learning}, and avoiding the use of modern deep modules (\textit{e.g.}, self-attention, dropout and batch/layer normalization) \cite{grathwohl2019your}, are adopted to mitigate the training instability issue caused by MCMC. These hard requirements limit the capacity of the deep model and might reduce its applicability to some large-scale datasets. More recently, an MCMC-free EBM training strategy \cite{grathwohl2021no} was proposed, where a generator is employed as a sampler to achieve amortized training of EBM. In particular, the output of the discriminator (also regarded as EBM) is a single scalar, hence, it is hard to directly estimate the entropy, resulting in complicated processing based on variational inference \cite{grathwohl2021no} or Jacobi-determinant \cite{geng2021bounds}. Inspired by the amortized training strategy, we generalize WGAN into a manifold representation and adopt a replay buffer strategy to directly estimate the entropy of the manifold representation. Compared with the state-of-the-art EBMs, the proposed method is simple but effective to estimate entropy, leading to stronger diversity and fidelity in image generation.  
 
\section{Method}

\begin{figure*}[!h]
    \centering
       \includegraphics[width=.88\textwidth]{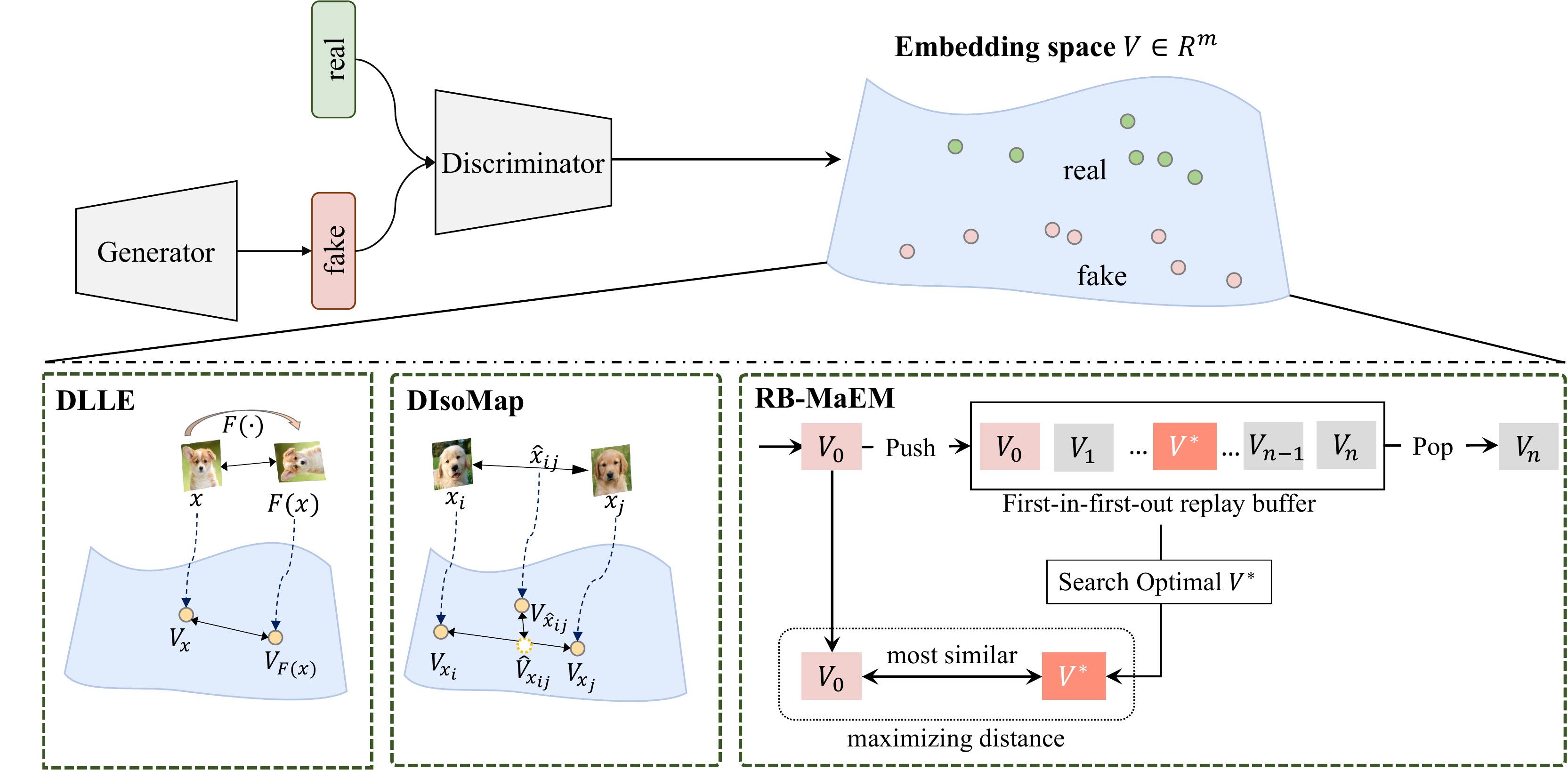}
   \caption{The pipeline of the proposed MaEM-GANs. Our discriminator embeds an input into an $m$-dimensional code $V$,  instead of a scalar value.   
   To  preserve the structural information of manifolds embedded in the input data, we introduce two regularization terms DLLE and DIsoMap. 
   Within the embedding space learned by the discriminator, RB-MaEM module is proposed to maximize the entropy of distributions  by maximizing the distance of most similar samples in the replay buffer.}
   \label{fig:pipeline}
\end{figure*}

The purpose of our method is to combat the mode collapse in GANs.
To this end, we first   bridge WGANs and  EBMs,  since EBMs have shown remarkable performance in avoiding the mode collapse. By analyzing the connections between WGANs and EBMs,  we discover that mode collapse in GANs can be prevented via maximizing the entropy of the generated distribution. 
However, it is nontrivial to directly estimate  the entropy of the generated distribution.
Instead, we propose a new training pipeline MaEM-GANs to approximate this objective.  In particular, we generalize the discriminator to feature embedding, to  embed images into a low-dimensional space. With the generalized discriminator, we propose a surrogate objective to maximize the entropy of the  distribution in the embedding space learned by the discriminator. RB-MaEM is proposed to efficiently optimize the surrogate objective.
Furthermore,  we introduce  two  regularization terms, \ie DLLE and  DIsoMap, to ensure that   the embedding space is well-formed and captures the underlying manifold embedded in the high-dimensional data.

\subsection{Problem Definition}
\label{sec:Problem}
Based on \cite{grathwohl2021no,geng2021bounds}, we revisit EBMs and WGANs. By analyzing the connections between them, we discover that the mode collapse can be prevented  by maximizing the entropy of  generated distribution.

\noindent
\textbf{Definition 1.}
An EBM can be represented from Gibbs density: 
\begin{align}
    p_\theta (x) = \frac{e^{f_\theta (x)}}{Z(\theta)}, 
\end{align}
where $f_\theta : \mathbb{R}^{h \times w} \rightarrow \mathbb{R}^1$ ($h$ and $w$ denote the height and width of the image, respectively) and $Z(\theta) = \int e^{f_\theta (x)} dx$ ($Z(\theta)$ is the partition function or normalizing constant).
EBMs are trained by maximum likelihood estimation:
\begin{align}
    \begin{split}
            \mathcal{L}_{ebm}(\theta)\! := \!-\!\!\!\!\underset{x\sim\mathcal{P}_r}{\mathbb{E}}[\text{log} (\frac{e^{f_\theta (x)}}{Z(\theta)})] \!=\! - \!\!\! \underset{x\sim\mathcal{P}_r}{\mathbb{E}}[f_\theta(x)] + \text{log} Z(\theta),
    \end{split}
\end{align}
where $\mathcal{P}_r$ is the data distribution.

\noindent
\textbf{Definition 2.}
The loss function of WGANs can be defined as
\begin{align}
\label{eq:wgan}
    \mathcal{L}_{wgan}(\theta) :=  \underset{x\sim\mathcal{P}_g}{\mathbb{E}}[\mathcal{D} (x)] - \underset{x\sim\mathcal{P}_r}{\mathbb{E}}[\mathcal{D}(x)],
\end{align}
where $\mathcal{P}_g$ is the generated distribution. Concretely, the discriminator $\mathcal{D}(\cdot)$ is trained by minimizing $\mathcal{L}_{wgan}(\theta)$, while the generator $\mathcal{G}(\cdot)$ is driven to maximize $\mathcal{L}_{wgan}(\theta)$.

\noindent
\textbf{Proposition 1.}
$\mathcal{L}_{wgan}(\theta)$ acts as a lower bound of $\mathcal{L}_{ebm}(\theta)$ by maximizing the entropy of the generated distribution $\mathcal{H}(\mathcal{P}_g)$. 

\noindent
\textbf{Proof:}
Given the probability density function $q(x)$ of $\mathcal{P}_g$ as input, an inequality can be derived as
\begin{align}
\label{eq:logz}
    \text{log} Z(\theta) \geq \text{log} Z(\theta) - \text{KL}(q(x)||p_\theta(x)),
\end{align}
where $\text{KL}(\cdot||\cdot)$ denotes the KL-divergence. 
With respect to $p_\theta (x) = e^{f_\theta (x)} / Z(\theta) $, $\text{log} Z(\theta)$ can be re-written as
\begin{align}
\begin{split}
        \text{log} Z(\theta)  &\geq \text{log}Z(\theta) + \int_x q(x) \text{log}(\frac{e^{f_\theta (x)}}{Z(\theta)q(x)}) \\
         &= \underset{x\sim\mathcal{P}_g}{\mathbb{E}}[f_\theta (x)] + H(\mathcal{P}_g),
\end{split}
\end{align}
where $\mathcal{H}(\mathcal{P}_g)$ is the entropy of $\mathcal{P}_g$. Note that model $f_\theta (x)$ in EBMs is equivalent to $\mathcal{D}(\cdot)$ in GANs. Hence, the connection between $\mathcal{L}_{wgan}$ and $\mathcal{L}_{ebm}$ can be established as 
\begin{align}
    \label{eq:bound}
    \begin{split}
      \mathcal{L}_{ebm}(\theta)  &= - \underset{x\sim\mathcal{P}_r}{\mathbb{E}}[f_\theta(x)] + \text{log} Z(\theta) \\
      &\geq \mathcal{L}_{wgan}(\theta) +  \mathcal{H}(\mathcal{P}_g)
    \end{split}
\end{align}

Based on Eq. (\ref{eq:bound}), $\mathcal{L}_{ebm}(\theta)$ is the upper bound of $\mathcal{L}_{wgan}(\theta)$  and an  entropy term $\mathcal{H}(\mathcal{P}_g)$. Generally, such an upper bound is very tight, because of the distance between $\mathcal{L}_{wgan}(\theta) + \mathcal{H}(\mathcal{P}_g)$ and $ \mathcal{L}_{ebm}(\theta)$ is the KL-divergence term $\text{KL}(q(x)||p_\theta(x))$ \cite{grathwohl2021no} (see Eq. (\ref{eq:logz})).
That is,  for the generator, we can  maximize  $\mathcal{L}_{wgan}$ and  the entropy term $\mathcal{H}(\mathcal{P}_g)$ to  approximate  $L_{ebm}$  that well alleviates the mode collapse issue. Intuitively, $\mathcal{H}(\mathcal{P}_g)$ plays a key role to tackle the mode collapse problem, which is generally neglected by existing GANs.


\subsection{Manifold Representation for WGANs}
As discussed above, mode collapse can be prevented by maximizing the entropy of the generated distribution. However, directly estimating the entropy $\mathcal{H}(\mathcal{P}_g)$ of the generated distribution is intractable, due to the large quantity and high dimensionality of data samples.
We address the above problem on the discriminator side,  motivated by the fact the discriminator heavily affects the training stability of GANs \cite{karras2020training}.
Specifically, we propose to generalize the discriminator from the perspective of the manifold and maximize the entropy of  distributions in the embedding space supported by the discriminator. 

Typical GANs \cite{arjovsky2017wasserstein,goodfellow2014generative,gulrajani2017improved,li2021anigan} treat the discriminator as a classifier, which classifies an image $x$ as real or fake according to a \textit{scalar} value $\mathcal{D}(x)$. 
Different from the scalar-based discriminators, our generalized discriminator $\mathbb{D}(\cdot)$ learns a mapping which transforms an input image to an $m$-dimensional embedding space: 
$\mathbb{D}(\cdot): \mathbf{R}^{h\times w} \rightarrow \mathbf{R}^{m}$,
where $h$ and $w$ are the height and weight of an image $x$, respectively, and $m \geq 1$.

The embedding space $\mathbb{D}(\cdot)$ is expected to capture informative characteristics of images. In other words, each dimension of the embedding space $\mathbb{D}(\cdot)$ corresponds to a critical attribute of images, such as color, texture, and structure. Thus, compared with the scalar-based discriminators, our generalized discriminator $\mathbb{D}$ can provide a more comprehensive representation for the data.

With our generalized discriminator $\mathbb{D}$, we reformulate the learning objective of WGANs in Eq. (\ref{eq:wgan}). Although the output of our  discriminator $\mathbb{D}$ is a vector, our discriminator determines whether an image $x$ is real/fake according to the average value of $\mathbb{D}(x)$. The new objective function $\mathcal{L}_{MaF}$ can be formulated as:
\begin{align}
       \mathcal{L}_{MaF} \triangleq 
       \underset{z \sim \mathcal{P}_{z}}{\mathbb{E}}[\frac{1}{m}\sum_k \mathbb{D}_k(\mathcal{G}(z))] - \underset{x \sim
       \mathcal{P}_r}{\mathbb{E}}[\frac{1}{m}\sum_{k}\mathbb{D}_k(x)],
\end{align}
where the standard normal distribution $\mathcal{N}(0,1)$ is used for $p_z$ and $\frac{1}{m}\sum_{k}\mathbb{D}_k(x)$ indicates the realness of $x$.

\subsection{Replay-Buffer-based Manifold Entropy Estimation}
With  our generalized discriminator $\mathbb{D}(\cdot)$, we design a surrogate objective to maximize the entropy of  distributions in the embedding space supported by $\mathbb{D}(\cdot)$. 

Given the $i^{th}$ image $x_i$, we represent it as an embedding code $V_i=\mathbb{D}(x_i)$ using our discriminator. With such representation, we aim to maximize  the entropy of  the embedding codes $\{V_i\}$'s distribution. 
Inspired by non-parametric entropy estimator \cite{beirlant1997nonparametric}, our insight is that if the distance between neighboring samples is maximized in a manifold, the data points in such a manifold will follow a uniform distribution, which maximizes the entropy of data distributions.
 We  hence propose  RB-MaEM module  to efficiently conduct this objective using a replay buffer. 
As shown in Fig. \ref{fig:pipeline},  RB-MaEM employs a first-in-first-out replay buffer $\mathcal{R}$  to store embedded codes $\{V_0,...,V_n\}$ of $n+1$   sample images, where $V_0$ is the code of the $0^{th}$ image  lying at the head of $\mathcal{R}$ and $V_n$ is the last code  at the tail of $\mathcal{R}$. We then search for an embedding code $V^*$ which is the most similar to $V_0$ from $V_1,...,V_n$ in buffer $\mathcal{R}$. With $V^*$, we maximize the entropy of the  distribution by maximizing  the distance of $V_0$ and its  most similar code $V^*$:
\begin{align}
     \label{eq:entropy}
     \mathcal{L}_{ent} =  \frac{V_0 \cdot V^* }{||V_0||\cdot ||V^* ||} + \lambda ||V_0||,
\end{align}
where $V^*=\arg\max_{V_i\in\mathcal{R}} (\frac{V_0 \cdot V_i}{||V_0||\, ||V_i||})$  obtained based on cosine similarity, and $\lambda ||V_0||$ is a regularization term to ensure the stabilization of the discriminator.

We further show that maximizing  the entropy of the generated distributions in the embedding space supported by   $\mathbb{D}(\cdot)$ approximately maximizes   the entropy $ \mathcal{H}(\mathcal{P}_g)$ of generated distribution which is stated in   \textbf{Proposition} 1.

\noindent
\textbf{Proposition 2.} The entropy of the distribution $\mathcal{P}_g$ in the embedding space learned by
the discriminator $\mathbb{D}(\cdot)$ is  the lower bound of $ \mathcal{H}(\mathcal{P}_g)$, based on the non-parametric entropy estimator and Lipschitz continuity.

\noindent
\textbf{Proof:} Based on the non-parametric entropy estimator \cite{beirlant1997nonparametric}, $ \mathcal{H}(\mathcal{P}_g)$ can be maximized by  
\begin{align}
\max \underset{x_i\sim\mathcal{P}_g}{\mathbb{E}}[\log(d_\mathcal{X} (x_i,x^*_i))], 
\label{eq:hp_proof2}
\end{align}
where $x^*_i \triangleq \underset{x_j \sim\mathcal{P}_g }{\arg\min} \, d_\mathcal{X}(x_i,x_j)$,
and $d_\mathcal{X}$ measures the distance between samples in image space $\mathcal{X}$. Based on the convergence of WGAN,  $\mathbb{D}$ is under Lipschitz continuity:
\begin{align}
     K  d_\mathcal{X}(x_i,x_j) \geq d_\mathbb{D}(\mathbb{D}(x_i),\mathbb{D}(x_j)),
     \label{eq:hp_lip_proof2}
\end{align}
where $K$ is the Lipschitz constant, and $d_\mathbb{D}$ is the metric in the embedding space supported by $\mathbb{D}$.  Since $K$ is typically set to be 1,  $d_\mathbb{D}(\mathbb{D}(x_i),\mathbb{D}(x_j))$  in the space of the discriminator can be regarded as the lower bound of the distance between $x_i$ and $x^*_i$ in image space. 
By using $d_\mathbb{D}(\mathbb{D}(x_i),\mathbb{D}(x_j))$ (\ie lower bound of $ d_\mathcal{X}(x_i,x_j)$ ) in  Eq. \ref {eq:hp_proof2} , we have 
\begin{align}
\begin{split}
    \max \!\!\! \underset{x_i\sim\mathcal{P}_g}{\mathbb{E}}&[\log(d_\mathcal{X} (x_i,x^*_i))] \geq \\ & \max\underset{x_i\sim\mathcal{P}_g}{\mathbb{E}}[\log(d_\mathbb{D} (\mathbb{D}(x_i),\mathbb{D}(x^*_i)))]
\end{split}
\end{align}
Therefore, the entropy of generated distribution in the image space can  be approximately maximized by maximizing the entropy in the embedding space.

\subsection{Manifold Regularization}
The performance of the proposed RB-MaEM depends on the quality of the embedding space learned by $\mathbb{D}(\cdot)$. To ensure that the  embedding space captures underlying manifolds embedded in the high-dimensional data, we introduce  DLLE and  DIsoMap to regularize the learning of $\mathbb{D}$.

\noindent
{\bf DLLE.} The first regularization term DLLE  enforces the discriminator $\mathbb{D}$ to preserve the nonlinear structure of high-dimensional data by using the local symmetries of linear reconstructions. In particular,  DLLE is established upon a simple geometric intuition \cite{roweis2000nonlinear}, \emph{i.e.}, a data sample and its neighbors lie on or are close to a locally linear region of the manifold learned by $\mathbb{D}(\cdot)$. Inspired by representation learning \cite{chen2020simple,zhang2019consistency},  we characterize  a different view of $x$ as the neighbor of $x$. Then, the learning objective $\mathcal{L}_{LLE}$ of DLLE enforces $x$ and its  neighbor to be similar in the embedding space of $\mathbb{D}(\cdot)$:
\begin{align}
   \mathcal{L}_{LLE} &\triangleq \underset{x \sim \mathcal{P}_r,\mathcal{P}_g}{\mathbb{E}}[||\mathbb{D}(\mathcal{F}(x)) - \mathbb{D}(x)||_2] \\
   &=\underset{x \sim \mathcal{P}_r,\mathcal{P}_g}{\mathbb{E}}[||V_{F(x)} - V_{x}||_2],
\end{align}
where $\mathcal{F}(\cdot)$ is the image transformation, such as rotation, adding Gaussian noise, and adversarial noise. By capturing the invariance between $x$  and its neighbors,   our DLLE helps the embedding space of $\mathbb{D}(\cdot)$   preserve the local geometry in the original data, leading to meaningful representations.

\noindent
{\bf DIsoMap.} 
Inspired by MaF-GANs~\cite{liu2021manifold}, we introduce DIsoMap to improve the embeddings' quality. In DIsoMap, we preserve the relationship between the embedding combination of  $\{\mathbb{D}(x_i),\mathbb{D}(x_j)\}$ and the embedding of the combination of $\{x_i,x_j\}$ with cosine similarity. Our DIsoMap further preserves the topological structure of different samples for manifold learning.

\section{Experimental Results and Analysis}
We implement our MaEM-GAN using the public PyTorch toolbox on eight NVIDIA V100 GPUs. To evaluate the performance of the proposed method, extensive experiments are carried on four publicly available datasets with different image sizes, including {\bf CIFAR-10} (32 $\times$ 32 pixels) \cite{krizhevsky2009learning},\footnote{https://www.cs.toronto.edu/~kriz/cifar.html} {\bf ANIMEFACE} (64$\times$64),\footnote{https://www.kaggle.com/splcher/animefacedataset} {\bf CelebA} (256 $\times$ 256) \cite{liu2015deep},\footnote{https://mmlab.ie.cuhk.edu.hk/projects/CelebA.html} and {\bf FFHQ} (1024 $\times$ 1024) \cite{karras2019style}.\footnote{https://github.com/NVlabs/ffhq-dataset} 
All the experimental settings, such as optimizer, network architecture and learning rate, are identical to the public benchmarks \cite{geng2021bounds,liu2021manifold,xiangli2020real,karras2019style}. Detailed information on the implementation of our MaEM-GAN can be found in {\itshape Supplementary Materials}.

\paragraph{\bf Evaluation metrics.} 
We use Fr\'{e}chet Inception Distance (FID) \cite{heusel2017gans} and Inception Score (IS) \cite{salimans2016improved}
to evaluate the quality of generated images.  Following state-of-the-art approaches \cite{geng2021bounds,xiangli2020real}, we use FID as our main evaluation  metric. 

To evaluate the effectiveness of our method on alleviating the mode collapse of GANs, we introduce $F_8$  and a new metric named \textit{I-Variance}, which measures the diversity of generated images to represent the degree of mode collapse. In other words, 
the larger diversity of the generated images indicates  a lower   degree of mode collapse. Hence,  I-Variance is defined as the standard deviation of generated distributions, where a generated image is represented by the extracted feature using Inception-V3 \cite{szegedy2016rethinking}:

\begin{align}
\text{I-Variance} \triangleq \sqrt{\underset{x\in \mathcal{P}_g}{\mathbb{E}}\big[||\mathcal{T}(x) - \underset{x\in \mathcal{P}_g}{\mathbb{E}}[\mathcal{T}(x)||_2 \big]},
\end{align}
where $\mathcal{T}(\cdot)$ is the Inception-V3 pre-trained on ImageNet. 
In all experiments, 50,000 images are randomly sampled to calculate FID, IS, and I-Variance.

\subsection{Ablation Studies and Analysis}%

\begin{table}[t]
\centering
\small
\begin{tabular}{>{\centering\arraybackslash}p{1.2cm}>{\centering\arraybackslash}p{1.2cm}>{\centering\arraybackslash}p{0.8cm}>{\centering\arraybackslash}p{1.8cm}|>{\centering\arraybackslash}p{1cm}}
\toprule
            Baseline        &   DIsoMap         &   DLLE            &   RB-MaEM     &   FID$\downarrow$                 \\ 
            \midrule
            $\checkmark$    &   $\times$        &   $\times$        &   $\times$        &   \textgreater{}100   \\
            $\checkmark$    &   $\checkmark$    &   $\times$        &   $\times$        &   51.96               \\
            $\checkmark$    &   $\checkmark$    &   $\checkmark$    &   $\times$        &   31.70               \\
            $\checkmark$    &   $\checkmark$    &   $\times$        &   $\checkmark$    &   30.73               \\
            \rowcolor{mygray}
            $\checkmark$    &   $\checkmark$    &   $\checkmark$    &   $\checkmark$    &   \textbf{29.22}      \\
            \bottomrule
\end{tabular}
\caption{The ablation study of the proposed method on CIFAR-10 in terms of FID.}
\label{tab:ablation}
\end{table}

\begin{table}[t]
\centering
\small
\begin{tabular}{>{\centering\arraybackslash}p{1.2cm}>{\centering\arraybackslash}p{1.5cm}>{\centering\arraybackslash}p{2.6cm}|>{\centering\arraybackslash}p{1cm}}
\toprule
w/ DLLE    &   Adv. Noise          &   Rot. and Gau. Noise             &   FID  $\downarrow$            \\
\midrule
$\checkmark$    &   $\times$            &   $\times$                        &   30.73           \\
$\checkmark$    &   $\checkmark$        &   $\times$                        &   33.04           \\
\rowcolor{mygray}
$\checkmark$    &   $\times$            &   $\checkmark$                    &   \textbf{29.22}  \\
\bottomrule
\end{tabular}
\caption{FID of the proposed method with different image transformation strategies $\mathcal{F}(\cdot)$ of DLLE on CIFAR-10. (Adv.--Adversarial; Rot.--Rotation; Gau.--Gaussian)}
\label{tab:dlle}
\end{table}

\noindent
{\bf Ablation studies.} To conduct ablation studies, we remove DIsoMap, DLLE, and  RB-MaEM  from our method to establish a baseline, \ie a high-dimensional WGAN with $\mathcal{L}_{MaF}$. Without our proposed components, the performance of this baseline is unsatisfactory in terms of FID score, as listed in Table \ref{tab:ablation}. More specifically, the discriminator and generator of the baseline with high dimension are trained in an imbalanced mode, which makes the training process unstable. In contrast, the performance of our method is significantly improved by adding  each proposed component to the baseline, validating the effectiveness of our components. As shown in Table \ref{tab:ablation}, DIsoMap is the key component to stabilize the training process, since this term can dynamically ensure the distance between the real and generated distributions to be tractable. Hence, DIsoMap drives the baseline to achieve an FID of 51.96. However, this term might lead to a trivial solution, \ie each dimension outputs the same value. To overcome the issue, DLLE and RB-MaEM are added and help the discriminator  yield non-trivial representation, which further improves our performance to 31.70 and 30.73 in FID, respectively.  
By incorporating all proposed components,  our method achieves the best FID score of 29.22. 

\begin{figure}[t]
    \centering
    \includegraphics[width=.45\textwidth]{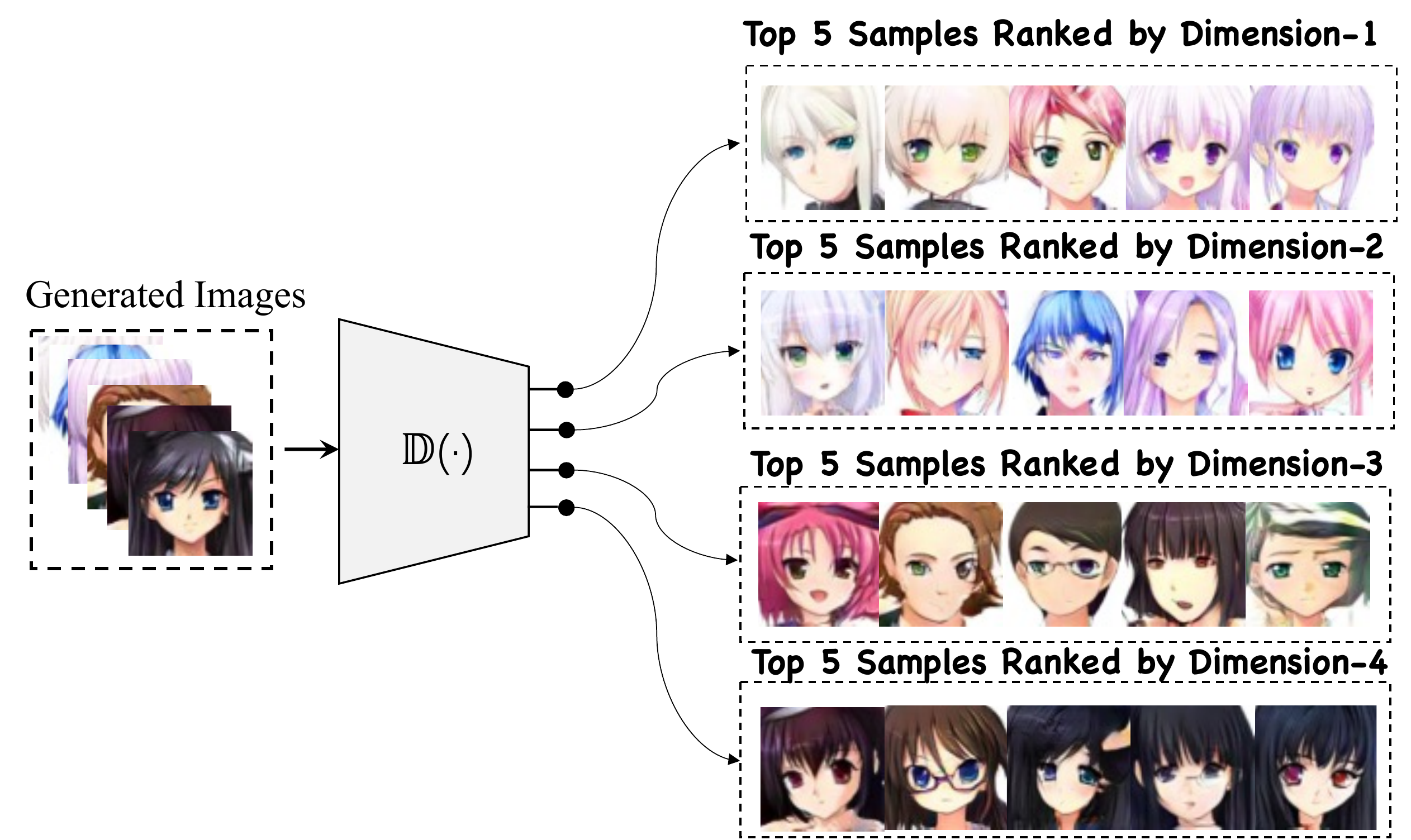}
   \caption{Top five generated images ranked by the individual dimension of the discriminator on ANIMEFACE.  In each row, a generated image is ranked  according to the $c^{\text{th}}$ dimension value of  its embeded  code $V$ extracted by  our discriminator ($c= 1,2,3,4$).}  
     \label{fig:outcome}
\end{figure}

\begin{table}[t]
\centering
\small

\begin{tabular}{>{\centering\arraybackslash}p{1.5cm}|>{\centering\arraybackslash}p{1cm}>{\centering\arraybackslash}p{1cm}>{\centering\arraybackslash}p{1cm}>{\centering\arraybackslash}p{1cm}>{\centering\arraybackslash}p{1cm}}

\toprule             
%
Buffer size&	128  & 512 &1024& 2048\\
\midrule
FID   	& 31.58  & 32.36 & \textbf{29.22}   & 31.87  \\
\bottomrule
\end{tabular}
\caption{ FID  $\downarrow$ of our method using different replay buffer sizes on CIFAR-10.}
\label{tab:buffer}
\end{table}

\begin{table}[t]
\centering
\small
\begin{tabular}{p{2.0cm}|>{\centering\arraybackslash}p{2.2cm}>{\centering\arraybackslash}p{2.5cm}}
\toprule
 Configuration                    & CIFAR-10 & ANIMEFACE                                  \\ 
 \midrule
 Baseline                   & 3.41 $\pm$ 0.03    & 2.46 $\pm$ 0.03                             \\ 
 \rowcolor{mygray}
 + RB-MaEM           & \textbf{4.88 $\pm$ 0.04}      & \textbf{2.59 $\pm$ 0.04}                           \\ 
 \bottomrule
\end{tabular}
\caption{The I-Variance score $\uparrow$ ($\times10^{-3}$) of our method on CIFAR-10 and ANIMEFACE. I-Variance score denotes the variance of the class-wise outputs from Inception-V3.}
\label{tab:variance}
\end{table}

\begin{table}[t]
\centering
\small
\begin{tabular}{>{\arraybackslash}p{4.7cm}|>{\centering\arraybackslash}p{1.5cm}>{\centering\arraybackslash}p{0.7cm}}

\toprule
Model         & CIFAR-10                        & CelebA                \\ 
\midrule
WGAN                            & 55.96                          &   -                    \\
HingeGAN                       & 42.40                          & 25.57                  \\
LSGAN                          & 42.01                           & 30.76                  \\
Std-GAN                         & 38,56                          & 27.02                  \\ 
WGAN-GP                        & 41.86                          & 70.28                  \\
\midrule
Realness GAN-Obj.1             & 36.73                          & -                      \\
Realness GAN-Obj.2             & 34.59                          & 23.51                  \\
Realness GAN-Obj.3             & 36.21                          & -                      \\ 
\midrule
MaF-CwGAN        & 39.24                          &          -              \\
MaF-DwGAN        & 33.73              &          -              \\
MaF-$\mathbb{E}$wGAN                & 30.85                 &     12.43                   \\ 
\rowcolor{mygray}
Ours                     & \textbf{29.22}                 &\textbf{9.14}  \\  
\bottomrule
\end{tabular}
\caption{FID $\downarrow$ of GAN models on CelebA and CIFAR-10. All methods use the same backbone of DCGAN \cite{radford2015unsupervised}. }
\label{tab:celeba}
\end{table}

\begin{table}[t]
\centering
\normalsize
\begin{tabular}{p{4.5cm}|>{\centering\arraybackslash}p{1cm}>{\centering\arraybackslash}p{1cm}}
\toprule
Model        & FID    $\downarrow$            & IS   $\uparrow$                                 \\ 
\midrule
Unconditional BigGAN           & 16.04                          & 9.10                                 \\
\rowcolor{mygray}
Unconditional BigGAN + Ours     & \textbf{13.86}                          & \textbf{9.27} \\ 
\bottomrule
\end{tabular}
\caption{FID  and Inception Score of BigGAN \cite{brock2018large} and our method on CIFAR-10.}
\label{tab:unconditionGAN}
\end{table}

\begin{table}[!hbpt]
\small
\centering
\begin{tabular}{p{6cm}|>{\centering\arraybackslash}p{1cm}}
\toprule
Model & FID $\downarrow$  \\
\midrule
StyleGAN-V1 \cite{karras2019style}  & 4.40 \\
StyleGAN-V2 \cite{karras2020analyzing} & 2.84 \\ 
\rowcolor{mygray}
StyleGAN-V2 + Ours & \textbf{2.67} \\ 
\bottomrule
\end{tabular}
\caption{FID  of  StyleGAN-V1/2 and our method on FFHQ.}
\label{tab:ffhq}
\end{table}

\begin{table}[!hbpt]
\centering
\small
\begin{tabular}{p{1.8cm}|>{\centering\arraybackslash}p{1.5cm}>{\centering\arraybackslash}p{1.5cm}>{\centering\arraybackslash}p{1.5cm}}
\toprule
Model             & IS   $\uparrow$               & FID   $\downarrow$    & $F_8$  $\uparrow$               \\ 
\midrule

MEG          & 2.20           & 9.31        & 0.95   \\ 
VERA               & 2.15           & 41.00       & 0.52   \\
EBM-0GP          & 2.26           & 20.53         & 0.89   \\ 
EBM-BB            & 2.26           & 12.75         & 0.94 \\ 
\midrule
DDPM$^*$     & 2.18                      & 8.81       & 0.94               \\ 
\midrule
WGAN-0GP       & 2.22           & 9.76  & 0.95         \\ 
\rowcolor{mygray}
Ours  & \textbf{2.80 } & \textbf{8.62 } & \textbf{0.98}\\ 
\bottomrule
\end{tabular}
\caption{FID, IS and $F_8$ of  EBMs, diffusion models and our method    on ANIMEFACE.}
\label{tab:ebm}
\end{table}

\noindent
{\bf Impact of Hyper-parameters.} For DLLE, Tab.  \ref{tab:dlle}  investigates different strategies of generating the neighbors of a given sample, where the first strategy adds the adversarial perturbation into the image.  The second strategy employs image augmentations \ie rotation and adding Gaussian noises, which is simpler yet achieves better performance in FID than the first one. 
In addition, Table \ref{tab:buffer} lists our results using different buffer sizes. Our method will be  degraded if the buff size is too small to store enough  meaningful samples for maximizing the entropy.  When the buffer size is 1024, our method achieves the best performance. 

\noindent
{\bf The effectiveness of the discriminator with manifold representation.}
In our method, the discriminator yields a vector to measure the realness of generated images. Each element in the vector measures the sample-wise realness from a specific attribute.
To demonstrate the effectiveness of manifold representation, we feed 50,000 generated images to our discriminator and obtain the corresponding vectors.
Fig.~\ref{fig:outcome} shows the `realest' samples ranked along different elements/dimensions of the obtained vectors.
We can observe that  the top samples ranked by different dimensions of our discriminator's output exhibit different attributes such as color and style.
Hence, introducing manifold representation  is effective to estimate the entropy and helps the discriminator to assess the realness from different aspects.

\noindent
{\bf The effectiveness of RB-MaEM on alleviating mode collapse.} We verify the importance of RB-MaEM in alleviating the mode collapse issue by measuring the diversity of generated images, where the I-Variance score is adopted to quantitatively assess the diversity of generated images. Table \ref{tab:variance} shows that our method without RB-MaEM only achieves  an I-Variance score of 3.41. In contrast, the I-Variance score increases to 4.88 by adding RB-MaEM. This shows that our RB-MaEM can significantly improve the diversity of generated images and effectively alleviate mode collapse.

\subsection{Comparison with State-of-the-art Methods}
To show the superiority of our method, we compare our method with recent state-of-the-art GAN models and energy-based models. Note that the results of all baseline methods are duplicated from the existing benchmarks \cite{xiangli2020real,geng2021bounds} without re-implementation.

\noindent
{\bf Comparison with GAN models.}  
Table~\ref{tab:celeba} shows the comparison results on CIFAR-10 and CelebA datasets. We compare our method with the state-of-the-art GANs, including WGAN \cite{arjovsky2017wasserstein}, HingeGAN \cite{zhao2016energy}, LSGAN  \cite{mao2017least}, Std-GAN \cite{goodfellow2014generative}, WGAN-GP \cite{gulrajani2017improved}, Realness GAN \cite{xiangli2020real}, and the family of MaF-GANs  \cite{liu2021manifold}.  
Compared with these methods, our method achieves the best FID score on both datasets.
Specifically, MaF-GAN is a recent work most related to ours, which also generalizes WGAN into a high-dimensional form. 
Our method surpasses MaF-GAN by a large margin since our method explicitly maximizes the entropy of distribution in the embedding space of the discriminator.
Furthermore, we demonstrate that  our method facilitates training  very deep GAN architectures even  with complicated training strategies. In particular, 
our method is incorporated into StyleGAN-V2 \cite{karras2020analyzing} and BigGAN \cite{brock2018large} by adding the proposed learning objective to their methods. And the consistent improvements can be found in Tables \ref{tab:celeba}, \ref{tab:ffhq} and \ref{tab:unconditionGAN}.

\noindent
{\bf Comparison with energy-based models.} 
We compare our method with representative energy-based models, including  MEG \cite{kumar2019maximum}, VERA  \cite{grathwohl2021no}, EBM-0GP \cite{geng2021bounds} and EBM-BB \cite{geng2021bounds}. 
For a more comprehensive study, we also include  Denoising Diffusion Probabilistic Model (DDPM) \cite{ho2020denoising} and WGAN-0GP \cite{thanh2019improving}. DDPM can be regarded as an upper bound for generation performance since this method generates images by expensive iterative optimization. 
Following \cite{geng2021bounds}, we conduct the experiments on the AnimeFace dataset. 
Table \ref{tab:ebm} shows that  our method outperforms  all energy-based models with the highest IS (2.80) and $F_8$ (0.98), and lowest FID (8.62),  demonstrating that our method not only ensures the fidelity of the generated images but also significantly improves  their diversity. 

\begin{figure}[!tb]
    \centering
    \includegraphics[width=0.46\textwidth]{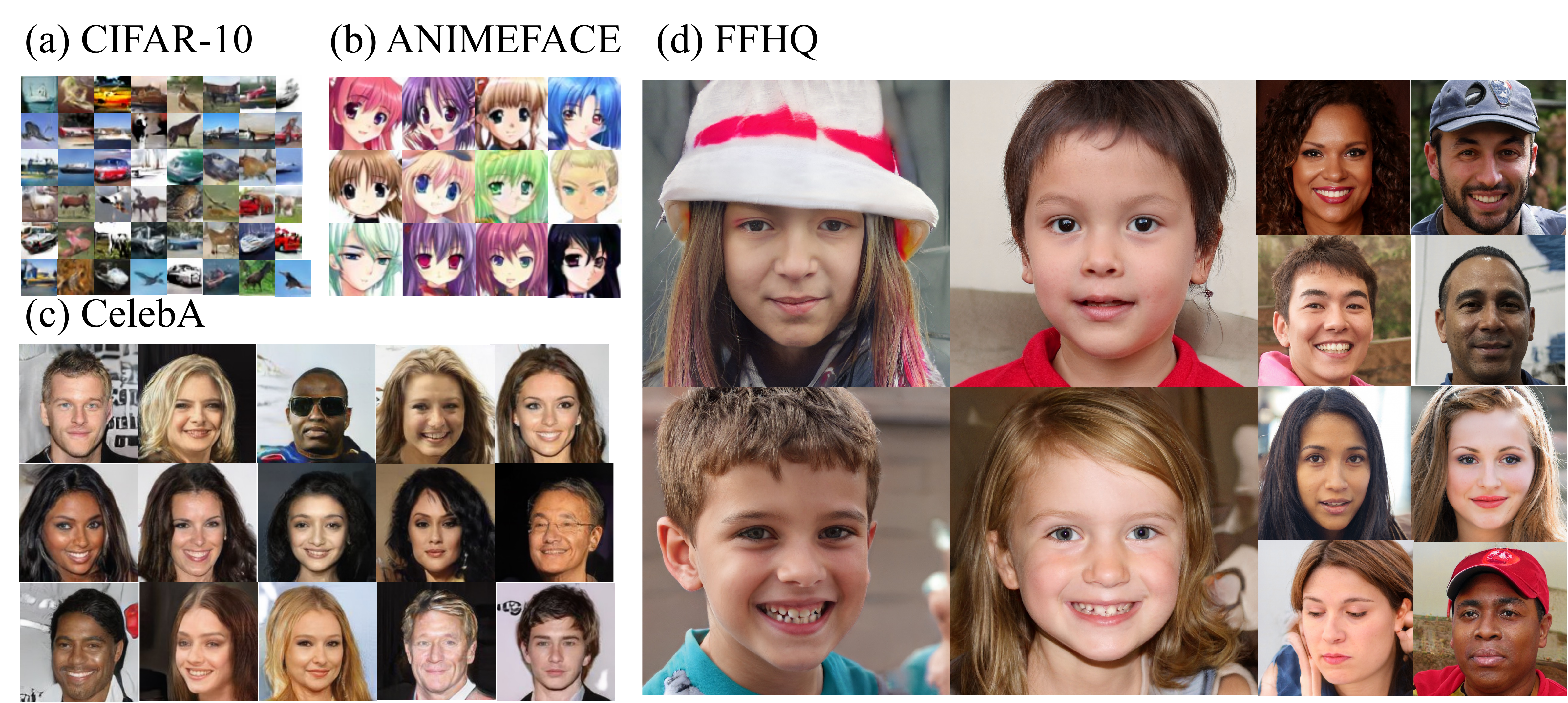}
   \caption{The generated samples of the proposed method on (a) CIFAR-10 (32 $\times$ 32), (b) ANIMEFACE (64 $\times$ 64), (c) CelebA (256 $\times$ 256) and (d) FFHQ (1024 $\times$ 1024).}
     \label{fig:ffhq}
\end{figure}

\section{Conclusion}
In this paper, we proposed a novel method to alleviate mode collapse in GANs. Our method generalizes the discriminator as a feature embedder, and mode collapse in GANs can be alleviated by maximizing the entropy of distributions in the embedding space learned by the discriminator. Two manifold regularization terms were introduced to preserve the information in the manifold embedded in the data. 
Based on the well-learned embedding space, a replay-buffer-based entropy estimator was proposed to maximize the diversity of samples in the embedding space.  By improving the
discriminator and maximizing the entropy of distributions in the
embedding space, our method effectively reduces the mode collapse
without sacrificing the quality of generated samples.  

\section{Acknowledgments}
This work was supported by the King Abdullah University of Science and Technology (KAUST) Office of Sponsored Research through the Visual Computing Center (VCC) funding, the Key-Area Research and Development Program of Guangdong Province, China (No. 2018B010111001), National Key R\&D Program of China (2018YFC2000702) and the Scientific and Technical Innovation 2030-"New Generation Artificial Intelligence" Project (No. 2020AAA0104100).  

\bibliography{egbib}

\begin{thebibliography}{51}
\providecommand{\natexlab}[1]{#1}

\bibitem[{An et~al.(2019)An, Guo, Lei, Luo, Yau, and Gu}]{an2019ae}
An, D.; Guo, Y.; Lei, N.; Luo, Z.; Yau, S.-T.; and Gu, X. 2019.
\newblock Ae-ot: a new generative model based on extended semi-discrete optimal
  transport.
\newblock In \emph{ICLR}.

\bibitem[{Arjovsky, Chintala, and Bottou(2017)}]{arjovsky2017wasserstein}
Arjovsky, M.; Chintala, S.; and Bottou, L. 2017.
\newblock Wasserstein generative adversarial networks.
\newblock In \emph{International Conference on Machine Learning}, 214--223.
  PMLR.

\bibitem[{AZhang et~al.(2020)AZhang, Zhang, Odena, and
  Lee}]{zhang2019consistency}
AZhang, H.; Zhang, Z.; Odena, A.; and Lee, H. 2020.
\newblock Consistency regularization for generative adversarial networks.
\newblock In \emph{ICLR}.

\bibitem[{Beirlant et~al.(1997)Beirlant, Dudewicz, Gy{\"o}rfi, Van~der Meulen
  et~al.}]{beirlant1997nonparametric}
Beirlant, J.; Dudewicz, E.~J.; Gy{\"o}rfi, L.; Van~der Meulen, E.~C.; et~al.
  1997.
\newblock Nonparametric entropy estimation: An overview.
\newblock \emph{International Journal of Mathematical and Statistical
  Sciences}, 6(1): 17--39.

\bibitem[{Brock, Donahue, and Simonyan(2019)}]{brock2018large}
Brock, A.; Donahue, J.; and Simonyan, K. 2019.
\newblock Large scale {GAN} training for high fidelity natural image synthesis.
\newblock \emph{ICLR}.

\bibitem[{Carlo(2004)}]{carlo2004markov}
Carlo, C.~M. 2004.
\newblock Markov {C}hain {M}onte {C}arlo and {G}ibbs sampling.
\newblock \emph{Lecture notes for EEB}, 581: 540.

\bibitem[{Chen et~al.(2020)Chen, Kornblith, Norouzi, and
  Hinton}]{chen2020simple}
Chen, T.; Kornblith, S.; Norouzi, M.; and Hinton, G. 2020.
\newblock A simple framework for contrastive learning of visual
  representations.
\newblock In \emph{International Conference on Machine Learning}, 1597--1607.
  PMLR.

\bibitem[{De~Sa, Chen, and Wong(2018)}]{de2018minibatch}
De~Sa, C.; Chen, V.; and Wong, W. 2018.
\newblock Minibatch {G}ibbs sampling on large graphical models.
\newblock In \emph{International Conference on Machine Learning}, 1173--1181.

\bibitem[{Du and Mordatch(2019)}]{du2019implicit}
Du, Y.; and Mordatch, I. 2019.
\newblock Implicit generation and modeling with energy based models.
\newblock \emph{Advances in Neural Information Processing Systems}, 32.

\bibitem[{Geng et~al.(2021)Geng, Wang, Gao, Frellsen, and
  Hauberg}]{geng2021bounds}
Geng, C.; Wang, J.; Gao, Z.; Frellsen, J.; and Hauberg, S. 2021.
\newblock Bounds all around: training energy-based models with bidirectional
  bounds.
\newblock \emph{Advances in Neural Information Processing Systems}, 34.

\bibitem[{Goodfellow et~al.(2014)Goodfellow, Pouget-Abadie, Mirza, Xu,
  Warde-Farley, Ozair, Courville, and Bengio}]{goodfellow2014generative}
Goodfellow, I.; Pouget-Abadie, J.; Mirza, M.; Xu, B.; Warde-Farley, D.; Ozair,
  S.; Courville, A.; and Bengio, Y. 2014.
\newblock Generative {A}dversarial {N}ets.
\newblock \emph{Advances in Neural Information Processing Systems}, 27.

\bibitem[{Goodfellow et~al.(2020)Goodfellow, Pouget-Abadie, Mirza, Xu,
  Warde-Farley, Ozair, Courville, and Bengio}]{goodfellow2020generative}
Goodfellow, I.; Pouget-Abadie, J.; Mirza, M.; Xu, B.; Warde-Farley, D.; Ozair,
  S.; Courville, A.; and Bengio, Y. 2020.
\newblock Generative adversarial networks.
\newblock \emph{Communications of the ACM}, 63(11): 139--144.

\bibitem[{Grathwohl et~al.(2020)Grathwohl, Wang, Jacobsen, Duvenaud, Norouzi,
  and Swersky}]{grathwohl2019your}
Grathwohl, W.; Wang, K.-C.; Jacobsen, J.-H.; Duvenaud, D.; Norouzi, M.; and
  Swersky, K. 2020.
\newblock Your classifier is secretly an energy based model and you should
  treat it like one.
\newblock \emph{International Conference on Learning Representations}.

\bibitem[{Grathwohl et~al.(2021)Grathwohl, Kelly, Hashemi, Norouzi, Swersky,
  and Duvenaud}]{grathwohl2021no}
Grathwohl, W.~S.; Kelly, J.~J.; Hashemi, M.; Norouzi, M.; Swersky, K.; and
  Duvenaud, D. 2021.
\newblock No {MCMC} for me: Amortized sampling for fast and stable training of
  energy-based models.
\newblock In \emph{International Conference on Learning Representations}.

\bibitem[{Gulrajani et~al.(2017)Gulrajani, Ahmed, Arjovsky, Dumoulin, and
  Courville}]{gulrajani2017improved}
Gulrajani, I.; Ahmed, F.; Arjovsky, M.; Dumoulin, V.; and Courville, A.~C.
  2017.
\newblock Improved training of {Wasserstein GANs}.
\newblock \emph{Advances in Neural Information Processing Systems}, 30.

\bibitem[{Heusel et~al.(2017)Heusel, Ramsauer, Unterthiner, Nessler, and
  Hochreiter}]{heusel2017gans}
Heusel, M.; Ramsauer, H.; Unterthiner, T.; Nessler, B.; and Hochreiter, S.
  2017.
\newblock {GANs} trained by a two time-scale update rule converge to a local
  {Nash} equilibrium.
\newblock \emph{Advances in Neural Information Processing Systems}, 30.

\bibitem[{Hinton and Sejnowski(1983)}]{hinton1983optimal}
Hinton, G.~E.; and Sejnowski, T.~J. 1983.
\newblock Optimal perceptual inference.
\newblock In \emph{Proceedings of the IEEE conference on Computer Vision and
  Pattern Recognition}, volume 448, 448--453. Citeseer.

\bibitem[{Ho, Jain, and Abbeel(2020)}]{ho2020denoising}
Ho, J.; Jain, A.; and Abbeel, P. 2020.
\newblock Denoising diffusion probabilistic models.
\newblock \emph{Advances in Neural Information Processing Systems}, 33:
  6840--6851.

\bibitem[{Hopfield(1982)}]{hopfield1982neural}
Hopfield, J.~J. 1982.
\newblock Neural networks and physical systems with emergent collective
  computational abilities.
\newblock \emph{Proceedings of the National Academy of Sciences}, 79(8):
  2554--2558.

\bibitem[{Karras et~al.(2020{\natexlab{a}})Karras, Aittala, Hellsten, Laine,
  Lehtinen, and Aila}]{karras2020training}
Karras, T.; Aittala, M.; Hellsten, J.; Laine, S.; Lehtinen, J.; and Aila, T.
  2020{\natexlab{a}}.
\newblock Training generative adversarial networks with limited data.
\newblock \emph{Advances in Neural Information Processing Systems}, 33:
  12104--12114.

\bibitem[{Karras, Laine, and Aila(2019)}]{karras2019style}
Karras, T.; Laine, S.; and Aila, T. 2019.
\newblock A style-based generator architecture for {Generative Adversarial
  Networks}.
\newblock In \emph{Proceedings of the IEEE/CVF Conference on Computer Vision
  and Pattern Recognition}, 4401--4410.

\bibitem[{Karras et~al.(2020{\natexlab{b}})Karras, Laine, Aittala, Hellsten,
  Lehtinen, and Aila}]{karras2020analyzing}
Karras, T.; Laine, S.; Aittala, M.; Hellsten, J.; Lehtinen, J.; and Aila, T.
  2020{\natexlab{b}}.
\newblock Analyzing and improving the image quality of {StyleGAN}.
\newblock In \emph{Proceedings of the IEEE/CVF Conference on Computer Vision
  and Pattern Recognition}, 8110--8119.

\bibitem[{Krizhevsky, Hinton et~al.(2009)}]{krizhevsky2009learning}
Krizhevsky, A.; Hinton, G.; et~al. 2009.
\newblock Learning multiple layers of features from tiny images.
\newblock \emph{Handbook of Systemic Autoimmune Diseases}, 1(4).

\bibitem[{Kumar et~al.(2019)Kumar, Ozair, Goyal, Courville, and
  Bengio}]{kumar2019maximum}
Kumar, R.; Ozair, S.; Goyal, A.; Courville, A.; and Bengio, Y. 2019.
\newblock Maximum entropy generators for energy-based models.
\newblock \emph{arXiv preprint arXiv:1901.08508}.

\bibitem[{LeCun et~al.(2006)LeCun, Chopra, Hadsell, Ranzato, and
  Huang}]{lecun2006tutorial}
LeCun, Y.; Chopra, S.; Hadsell, R.; Ranzato, M.; and Huang, F. 2006.
\newblock A tutorial on energy-based learning.
\newblock \emph{Predicting structured data}, 1(0).

\bibitem[{Li et~al.(2021)Li, Zhu, Wang, Lin, Ghanem, and Shen}]{li2021anigan}
Li, B.; Zhu, Y.; Wang, Y.; Lin, C.-W.; Ghanem, B.; and Shen, L. 2021.
\newblock AniGAN: Style-Guided Generative Adversarial Networks for Unsupervised
  Anime Face Generation.
\newblock \emph{IEEE Transactions on Multimedia}.

\bibitem[{Li et~al.(2019)Li, Yang, Liu, Yang, Jeon, and Wu}]{li2019feedback}
Li, Z.; Yang, J.; Liu, Z.; Yang, X.; Jeon, G.; and Wu, W. 2019.
\newblock Feedback network for image super-resolution.
\newblock In \emph{Proceedings of the IEEE/CVF conference on computer vision
  and pattern recognition}, 3867--3876.

\bibitem[{Liu et~al.(2021)Liu, Liang, Hou, Wu, Liu, and Shen}]{liu2021manifold}
Liu, H.; Liang, H.; Hou, X.; Wu, H.; Liu, F.; and Shen, L. 2021.
\newblock Manifold-preserved {GANs}.
\newblock \emph{arXiv preprint arXiv:2109.08955}.

\bibitem[{Liu et~al.(2015)Liu, Luo, Wang, and Tang}]{liu2015deep}
Liu, Z.; Luo, P.; Wang, X.; and Tang, X. 2015.
\newblock Deep learning face attributes in the wild.
\newblock In \emph{Proceedings of the IEEE International Conference on Computer
  Vision}, 3730--3738.

\bibitem[{Mangalam and Garg(2021)}]{mangalam2021overcoming}
Mangalam, K.; and Garg, R. 2021.
\newblock Overcoming Mode Collapse with Adaptive Multi Adversarial Training.
\newblock \emph{arXiv preprint arXiv:2112.14406}.

\bibitem[{Mao et~al.(2017)Mao, Li, Xie, Lau, Wang, and
  Paul~Smolley}]{mao2017least}
Mao, X.; Li, Q.; Xie, H.; Lau, R.~Y.; Wang, Z.; and Paul~Smolley, S. 2017.
\newblock Least squares generative adversarial networks.
\newblock In \emph{Proceedings of the IEEE International Conference on Computer
  Vision}, 2794--2802.

\bibitem[{Mirza and Osindero(2014)}]{mirza2014conditional}
Mirza, M.; and Osindero, S. 2014.
\newblock Conditional {Generative} {Adversarial} {N}ets.
\newblock \emph{arXiv preprint arXiv:1411.1784}.

\bibitem[{Nijkamp et~al.(2019)Nijkamp, Hill, Zhu, and Wu}]{nijkamp2019learning}
Nijkamp, E.; Hill, M.; Zhu, S.-C.; and Wu, Y.~N. 2019.
\newblock Learning non-convergent non-persistent short-run {MCMC} toward
  energy-based model.
\newblock \emph{Advances in Neural Information Processing Systems}, 32.

\bibitem[{Odena, Olah, and Shlens(2017)}]{odena2017conditional}
Odena, A.; Olah, C.; and Shlens, J. 2017.
\newblock Conditional image synthesis with auxiliary classifier {GANs}.
\newblock In \emph{International Conference on Machine Learning}, 2642--2651.
  PMLR.

\bibitem[{Radford, Metz, and Chintala(2016)}]{radford2015unsupervised}
Radford, A.; Metz, L.; and Chintala, S. 2016.
\newblock Unsupervised representation learning with deep convolutional
  generative adversarial networks.
\newblock \emph{International Conference on Learning Representations}.

\bibitem[{Roweis and Saul(2000)}]{roweis2000nonlinear}
Roweis, S.~T.; and Saul, L.~K. 2000.
\newblock Nonlinear dimensionality reduction by locally linear embedding.
\newblock \emph{science}, 290(5500): 2323--2326.

\bibitem[{Salimans et~al.(2016)Salimans, Goodfellow, Zaremba, Cheung, Radford,
  and Chen}]{salimans2016improved}
Salimans, T.; Goodfellow, I.; Zaremba, W.; Cheung, V.; Radford, A.; and Chen,
  X. 2016.
\newblock Improved techniques for training {GANs}.
\newblock \emph{Advances in Neural Information Processing Systems}, 29.

\bibitem[{Saxena and Cao(2021)}]{saxena2021generative}
Saxena, D.; and Cao, J. 2021.
\newblock Generative adversarial networks ({GANs}) challenges, solutions, and
  future directions.
\newblock \emph{ACM Computing Surveys (CSUR)}, 54(3): 1--42.

\bibitem[{Schmidhuber(1990)}]{schmidhuber1990making}
Schmidhuber, J. 1990.
\newblock Making the world differentiable: On using fully recurrent
  self-supervised neural networks for dynamic reinforcement learning and
  planning in non-stationary environments.
\newblock \emph{Institut f{\"u}r Informatik, Technische Universit{\"a}t
  M{\"u}nchen. Technical Report FKI-126}, 90.

\bibitem[{Schmidhuber(1991)}]{schmidhuber1991possibility}
Schmidhuber, J. 1991.
\newblock A possibility for implementing curiosity and boredom in
  model-building neural controllers.
\newblock In \emph{Proc. of the International Conference on Simulation of
  Adaptive Behavior: From Animals to Animats}, 222--227.

\bibitem[{Schmidhuber(2020)}]{schmidhuber2020generative}
Schmidhuber, J. 2020.
\newblock Generative adversarial networks are special cases of artificial
  curiosity (1990) and also closely related to predictability minimization
  (1991).
\newblock \emph{Neural Networks}, 127: 58--66.

\bibitem[{Szegedy et~al.(2016)Szegedy, Vanhoucke, Ioffe, Shlens, and
  Wojna}]{szegedy2016rethinking}
Szegedy, C.; Vanhoucke, V.; Ioffe, S.; Shlens, J.; and Wojna, Z. 2016.
\newblock Rethinking the {I}nception architecture for computer vision.
\newblock In \emph{Proceedings of the IEEE Conference on Computer Vision and
  Pattern Recognition}, 2818--2826.

\bibitem[{Thanh-Tung, Tran, and Venkatesh(2019)}]{thanh2019improving}
Thanh-Tung, H.; Tran, T.; and Venkatesh, S. 2019.
\newblock Improving generalization and stability of generative adversarial
  networks.
\newblock \emph{International Conference on Learning Representations}.

\bibitem[{Xiangli et~al.(2020)Xiangli, Deng, Dai, Loy, and
  Lin}]{xiangli2020real}
Xiangli, Y.; Deng, Y.; Dai, B.; Loy, C.~C.; and Lin, D. 2020.
\newblock Real or not real, that is the question.
\newblock \emph{International Conference on Learning Representations}.

\bibitem[{Xie et~al.(2018{\natexlab{a}})Xie, Lu, Gao, Zhu, and
  Wu}]{xie2018cooperative}
Xie, J.; Lu, Y.; Gao, R.; Zhu, S.-C.; and Wu, Y.~N. 2018{\natexlab{a}}.
\newblock Cooperative training of descriptor and generator networks.
\newblock \emph{IEEE transactions on pattern analysis and machine
  intelligence}, 42(1): 27--45.

\bibitem[{Xie et~al.(2016)Xie, Lu, Zhu, and Wu}]{xie2016theory}
Xie, J.; Lu, Y.; Zhu, S.-C.; and Wu, Y. 2016.
\newblock A theory of generative convnet.
\newblock In \emph{International Conference on Machine Learning}, 2635--2644.
  PMLR.

\bibitem[{Xie et~al.(2018{\natexlab{b}})Xie, Zheng, Gao, Wang, Zhu, and
  Wu}]{xie2018learning}
Xie, J.; Zheng, Z.; Gao, R.; Wang, W.; Zhu, S.-C.; and Wu, Y.~N.
  2018{\natexlab{b}}.
\newblock Learning descriptor networks for 3d shape synthesis and analysis.
\newblock In \emph{Proceedings of the IEEE conference on computer vision and
  pattern recognition}, 8629--8638.

\bibitem[{Xie, Zhu, and Nian~Wu(2017)}]{xie2017synthesizing}
Xie, J.; Zhu, S.-C.; and Nian~Wu, Y. 2017.
\newblock Synthesizing dynamic patterns by spatial-temporal generative convnet.
\newblock In \emph{Proceedings of the IEEE conference on computer vision and
  pattern recognition}, 7093--7101.

\bibitem[{Xie, Zhu, and Wu(2019)}]{xie2019learning}
Xie, J.; Zhu, S.-C.; and Wu, Y.~N. 2019.
\newblock Learning energy-based spatial-temporal generative convnets for
  dynamic patterns.
\newblock \emph{IEEE transactions on pattern analysis and machine
  intelligence}, 43(2): 516--531.

\bibitem[{Yu et~al.(2018)Yu, Lin, Yang, Shen, Lu, and Huang}]{yu2018generative}
Yu, J.; Lin, Z.; Yang, J.; Shen, X.; Lu, X.; and Huang, T.~S. 2018.
\newblock Generative image inpainting with contextual attention.
\newblock In \emph{Proceedings of the IEEE conference on computer vision and
  pattern recognition}, 5505--5514.

\bibitem[{Zhao, Mathieu, and LeCun(2017)}]{zhao2016energy}
Zhao, J.; Mathieu, M.; and LeCun, Y. 2017.
\newblock Energy-based generative adversarial network.
\newblock \emph{International Conference on Learning Representations}.

\end{thebibliography}

\end{document}